\documentclass{article}
\usepackage{nips99e}
\usepackage{epsf}
\newtheorem{theorem}{Theorem}

\def\lforall#1{\forall \: #1 \;}

\def\comment#1{}

\def\mtuple#1{\langle #1\rangle}

\def\centerps#1{\begin{center}
\leavevmode
\epsfbox{#1}
\end{center}}

\def\condition#1{{\noindent \bf #1}}

\title{State Abstraction in MAXQ Hierarchical Reinforcement Learning}

\author{Thomas G.~Dietterich \\
Department of Computer Science\\ 
Oregon State University\\
Corvallis, Oregon 97331-3202\\
{\it tgd@cs.orst.edu}}

\begin{document}

\maketitle

\begin{abstract}
Many researchers have explored methods for hierarchical reinforcement
learning (RL) with {\it temporal abstractions}, in which abstract
actions are defined that can perform many primitive actions before
terminating.  However, little is known about learning with {\it state
abstractions,} in which aspects of the state space are ignored.  In
previous work, we developed the MAXQ method for hierarchical RL.  In
this paper, we define five conditions under which state abstraction
can be combined with the MAXQ value function decomposition.  We prove
that the MAXQ-Q learning algorithm converges under these conditions
and show experimentally that state abstraction is important for the
successful application of MAXQ-Q learning.\\
{\bf Category:} Reinforcement Learning and Control\\
{\bf Preference:} Oral
\end{abstract}

\section{Introduction} 

Most work on hierarchical reinforcement learning has focused on
temporal abstraction.  For example, in the Options framework
\cite{nips-10:Precup+Sutton:1998,sps-bmsm:lprkmts-98}, the programmer
defines a set of macro actions (``options'') and provides a policy for
each.  Learning algorithms (such as semi-Markov Q learning) can then
treat these temporally abstract actions as if they were primitives and
learn a policy for selecting among them.  Closely related is the HAM
framework, in which the programmer constructs a hierarchy of
finite-state controllers \cite{nips-10:Parr+Russell:1998}.  Each
controller can include non-deterministic states (where the programmer
was not sure what action to perform).  The HAMQ learning algorithm can
then be applied to learn a policy for making choices in the
non-deterministic states.  In both of these approaches---and in other
studies of hierarchical RL (e.g.,
\cite{mach:Singh:1992,k-hrl:pr-93,hmbkd-hsmdpm-98})---each option or
finite state controller must have access to the entire state space.
The one exception to this---the Feudal-Q method of Dayan and Hinton
\cite{dh-frl-93}---introduced state abstractions in an unsafe way,
such that the resulting learning problem was only partially
observable.  Hence, they could not provide any formal results
for the convergence or performance of their method.

Even a brief consideration of human-level intelligence shows that such
methods cannot scale.  When deciding how to walk from the bedroom to
the kitchen, we do not need to think about the location of our car.
Without state abstractions, any RL method that learns value functions
must learn a separate value for each state of the world.  Some argue
that this can be solved by clever value function approximation
methods---and there is some merit in this view.  In this paper,
however, we explore a different approach in which we identify aspects
of the MDP that permit state abstractions to be safely incorporated in
a hierarchical reinforcement learning method without introducing
function approximations.  This permits us to obtain the first proof of
the convergence of hierarchical RL to an optimal policy in the
presence of state abstraction.

We introduce these state abstractions within the MAXQ framework
\cite{d-mmhrl-98}, but the basic ideas are general.  In our previous
work with MAXQ, we briefly discussed state abstractions, and we
employed them in our experiments.  However, we could not prove that
our algorithm (MAXQ-Q) converged with state abstractions, and we did
not have a usable characterization of the situations in which state
abstraction could be safely employed.  This paper solves these
problems and in addition compares the effectiveness of MAXQ-Q learning
with and without state abstractions.  The results show that state
abstraction is very important, and in most cases essential, to the
effective application of MAXQ-Q learning.

\section{The MAXQ Framework}

Let $M$ be a Markov decision problem with states $S$, actions $A$,
reward function $R(s'|s,a)$ and probability transition function
$P(s'|s,a)$.  Our results apply in both the finite-horizon
undiscounted case and the infinite-horizon discounted case.  Let
$\{M_0, \ldots, M_n\}$ be a set of subtasks of $M$, where each subtask
$M_i$ is defined by a termination predicate $T_i$ and a set of actions
$A_i$ (which may be other subtasks or primitive actions from $A$).
The ``goal'' of subtask $M_i$ is to move the environment into a state
such that $T_i$ is satisfied.  (This can be refined using a local
reward function to express preferences among the different states
satisfying $T_i$ \cite{d-mmhrl-98}, but we omit this refinement in
this paper.)  The subtasks of $M$ must form a DAG with a single
``root'' node---no subtask may invoke itself directly or indirectly.
A {\it hierarchical policy} is a set of policies $\pi = \{\pi_0,
\ldots, \pi_n\}$, one for each subtask.  A hierarchical policy is
executed using standard procedure-call-and-return semantics, starting
with the root task $M_0$ and unfolding recursively until primitive
actions are executed.  When the policy for $M_i$ is invoked in state
$s$, let $P(s',N|s,i)$ be the probability that it terminates in state
$s'$ after executing $N$ primitive actions.  A hierarchical policy is
{\it recursively optimal} if each policy $\pi_i$ is optimal given the
policies of its descendants in the DAG.

Let $V(i,s)$ be the value function for subtask $i$ in state $s$ (i.e.,
the value of following some policy starting in $s$ until we reach a
state $s'$ satisfying $T_i(s')$).  Similarly, let $Q(i,s,j)$ be the
$Q$ value for subtask $i$ of executing child action $j$ in state $s$
and then executing the current policy until termination.  The MAXQ
value function decomposition is based on the observation that each
subtask $M_i$ can be viewed as a Semi-Markov Decision problem in which
the reward for performing action $j$ in state $s$ is equal to
$V(j,s)$, the value function for subtask $j$ in state $s$.  To see
this, consider the sequence of rewards $r_t$ that will be received
when we execute child action $j$ and then continue with subsequent
actions according to hierarchical policy $\pi$:
\[
Q(i,s,j) = E\{r_t + \gamma r_{t+1} + \gamma^2 r_{t+2} + \cdots | s_t = s, \pi\}
\]
The macro action $j$ will execute for some number of steps $N$ and
then return.  Hence, we can partition this sum into two terms:
\[
Q(i,s,j) = E\left\{\left.\sum_{u=0}^{N-1} \gamma^u r_{t+u}\;  +
\;\sum_{u=N}^{\infty} \gamma^u r_{t+u}
\right| s_t = s, \pi\right\}
\]
The first term is the discounted sum of rewards until subtask $j$
terminates---$V(j,s)$.  The second term is the cost of finishing
subtask $i$ {\it after} $j$ is executed (discounted to the time when
$j$ is initiated).  We call this second term the {\it completion
function}, and denote it $C(i,s,j)$.  We can then write the Bellman
equation as
\begin{eqnarray*}
Q(i,s,j) &=& \sum_{s',N} P(s',N|s,j) \cdot  [V(j,s) + \gamma^N \max_{j'} Q(i,s',j')]\\
         &=& V(j,s) + C(i,s,j)
\end{eqnarray*}
To terminate this recursion, define $V(a,s)$ for a primitive action
$a$ to be the expected reward of performing action $a$ in state $s$. 

The MAXQ-Q learning algorithm is a simple variation of $Q$ learning in
which at subtask $M_i$, state $s$, we choose a child action $j$ and
invoke its (current) policy.  When it returns, we observe the
resulting state $s'$ and the number of elapsed time steps $N$ and
update $C(i,s,j)$ according to
\[C(i,s,j) := (1-\alpha_t)C(i,s,j) + \alpha_t \cdot \gamma^N [\max_{a'}
V(a',s') + C(i,s',a')].\]

To prove convergence, we require that the exploration policy executed
during learning be an {\it ordered GLIE policy}.  An {\it ordered
policy} is a policy that breaks Q-value ties among actions by
preferring the action that comes first in some fixed ordering.  A {\it
GLIE policy} \cite{sjls-crssoprla-98} is a policy that (a) executes
each action infinitely often in every state that is visited infinitely
often and (b) converges with probability 1 to a greedy policy.  The
ordering condition is required to ensure that the recursively optimal
policy is unique.  Without this condition, there are potentially many
different recursively optimal policies {\it with different values},
depending on how ties are broken within subtasks, subsubtasks, and so
on.

\begin{theorem} \label{theorem-converge}
Let $M = \mtuple{S, A, P, R}$ be either an episodic MDP for which all
deterministic policies are proper or a discounted infinite horizon MDP
with discount factor $\gamma$.  Let $H$ be a DAG defined over subtasks
$\{M_0, \ldots, M_k\}$.  Let $\alpha_t(i) > 0$ be a sequence of
constants for each subtask $M_i$ such that
\begin{equation}
\lim_{T\rightarrow \infty} \sum_{t=1}^T \alpha_t(i) = \infty \;\;\;\;
\mbox{and} \;\;\;\;
\lim_{T\rightarrow \infty} \sum_{t=1}^T \alpha_t^2(i) < \infty 
\label{alpha-diverge-converge}
\end{equation}
Let $\pi_x(i,s)$ be an ordered GLIE policy at each subtask $M_i$ and state
$s$ and assume that $|V_t(i,s)|$ and $|C_t(i,s,a)|$ are bounded for
all $t$, $i$, $s$, and $a$.  Then with probability 1, algorithm MAXQ-Q
converges to the unique recursively optimal policy for $M$ consistent
with $H$ and $\pi_x$.
\end{theorem}
\condition{Proof: (sketch)}
The proof is based on Proposition 4.5 from Bertsekas and Tsitsiklis
\cite{bt-ndp-96} and follows the standard stochastic approximation
argument due to \cite{nc:Jaakkola+Jordan+Singh:1994} generalized to
the case of non-stationary noise.  There are two key points in the
proof.  Define $P_t(s',N|s,j)$ to be the probability transition
function that describes the behavior of executing the current policy
for subtask $j$ at time $t$.  By an inductive argument, we show that
this probability transition function converges (w.p.~1) to the
probability transition function of the recursively optimal policy for
$j$.  Second, we show how to convert the usual weighted max norm
contraction for $Q$ into a weighted max norm contraction for $C$.
This is straightforward, and completes the proof.

What is notable about MAXQ-Q is that it can learn the value functions
of all subtasks simultaneously---it does not need to wait for the
value function for subtask $j$ to converge before beginning to learn
the value function for its parent task $i$.  This gives a completely
online learning algorithm with wide applicability.

\section{Conditions for Safe State Abstraction}

To motivate state abstraction, consider the simple Taxi Task shown in
Figure~\ref{fig-taxi}.  There are four special locations in this
world, marked as R(ed), B(lue), G(reen), and Y(ellow).  In each
episode, the taxi starts in a randomly-chosen square.  There is a
passenger at one of the four locations (chosen randomly), and that
passenger wishes to be transported to one of the four locations (also
chosen randomly).  The taxi must go to the passenger's location (the
``source''), pick up the passenger, go to the destination location
(the ``destination''), and put down the passenger there.  The episode
ends when the passenger is deposited at the destination location.

{\footnotesize
\def\normalsize{\footnotesize}
\begin{figure}
\begin{minipage}{1.5in}
\centerps{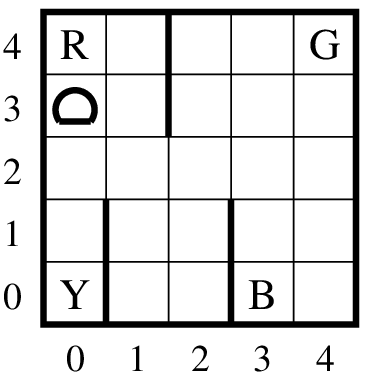}
\end{minipage}
\hfil
\begin{minipage}{3.4in}
{\epsfxsize=3.4in
\centerps{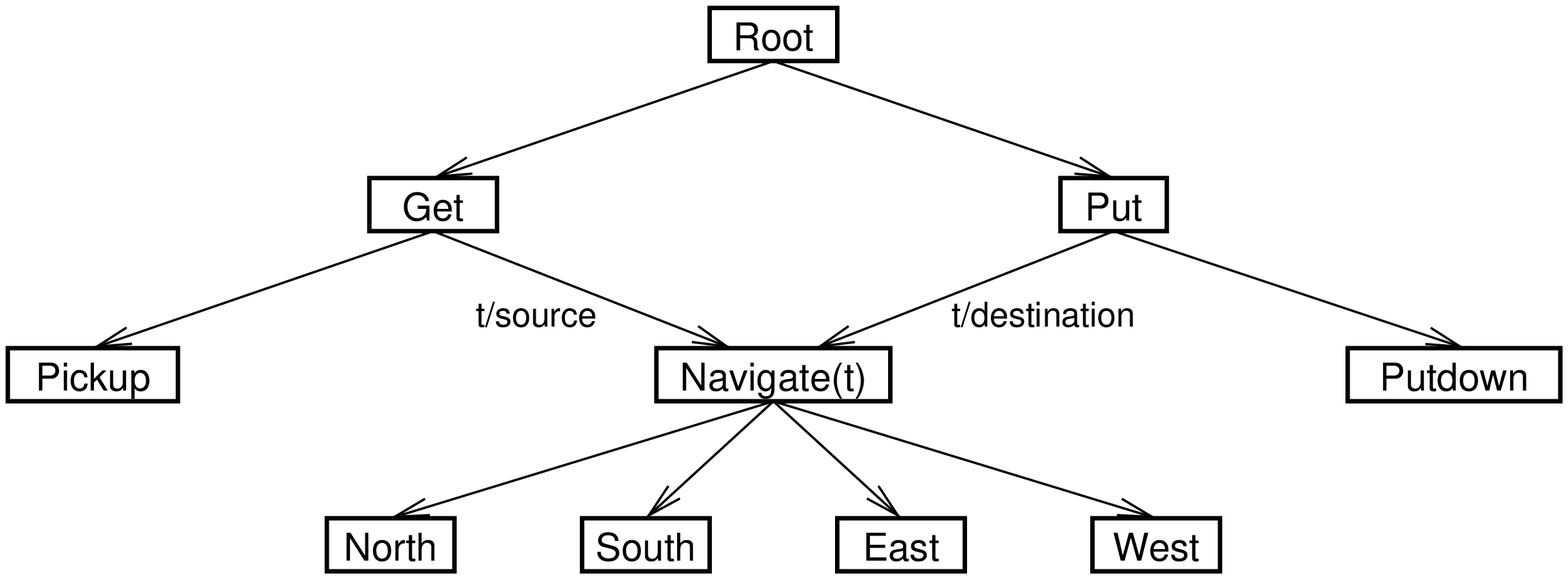}
}
\end{minipage}
\caption{Left: The Taxi Domain (taxi at row 3 column 0). Right: Task
Graph.}
\label{fig-taxi}
\end{figure}
}

There are six primitive actions in this domain: (a) four navigation
actions that move the taxi one square {\sf North}, {\sf South}, {\sf
East}, or {\sf West}, (b) a {\sf Pickup} action, and (c) a {\sf
Putdown} action.  Each action is deterministic.  There is a reward of
$-1$ for each action and an additional reward of $+20$ for
successfully delivering the passenger.  There is a reward of $-10$ if
the taxi attempts to execute the {\sf Putdown} or {\sf Pickup} actions
illegally.  If a navigation action would cause the taxi to hit a wall,
the action is a no-op, and there is only the usual reward of $-1$.

This task has a hierarchical structure (see Fig.~\ref{fig-taxi}) in
which there are two main sub-tasks: Get the passenger ({\sf Get}) and
Deliver the passenger ({\sf Put}).  Each of these subtasks in turn
involves the subtask of navigating to one of the four locations (${\sf
Navigate}(t)$; where $t$ is bound to the desired target location) and
then performing a {\sf Pickup} or {\sf Putdown} action.  This task
illustrates the need to support both temporal abstraction and state
abstraction.  The temporal abstraction is obvious---for example, {\sf
Get} is a temporally extended action that can take different numbers
of steps to complete depending on the distance to the target.  The top
level policy (get passenger; deliver passenger) can be expressed very
simply with these abstractions.

The need for state abstraction is perhaps less obvious.  Consider the
{\sf Get} subtask.  While this subtask is being solved, the
destination of the passenger is completely irrelevant---it cannot
affect any of the nagivation or pickup decisions.  Perhaps more
importantly, when navigating to a target location (either the source
or destination location of the passenger), only the taxi's location
and identity of the target location are important.  The fact that in
some cases the taxi is carrying the passenger and in other cases it is
not is irrelevant.

We now introduce the five conditions for state abstraction.  We will
assume that the state $s$ of the MDP is represented as a vector of
state variables.  A state abstraction can be defined for each
combination of subtask $M_i$ and child action $j$ by identifying a
subset $X$ of the state variables that are {\it relevant} and defining
the value function and the policy using only these relevant variables.
Such value functions and policies are said to be {\it abstract}.

The first two conditions involve eliminating irrelevant variables
within a subtask of the MAXQ decomposition.

\condition{Condition 1: Subtask Irrelevance.}  Let $M_i$ be a subtask
of MDP $M$.  A set of state variables $Y$ is {\em irrelevant to
subtask $i$} if the state variables of $M$ can be partitioned into two
sets $X$ and $Y$ such that for any stationary abstract hierarchical
policy $\pi$ executed by the descendants of $M_i$, the following two
properties hold: (a) the state transition probability distribution
$P^{\pi}(s',N|s,j)$ for each child action $j$ of $M_i$ can be factored
into the product of two distributions:
\begin{equation} \label{eq-p-independence}
P^{\pi}(x',y',N|x,y,j) = P^{\pi}(x',N|x,j) \cdot P^{\pi}(y'|y,j),
\end{equation}
where $x$ and $x'$ give values for the variables in $X$, and $y$ and
$y'$ give values for the variables in $Y$; and (b) for any pair
of states $s_1=(x,y_1)$ and $s_2=(x,y_2)$ and any child action $j$,
$V^{\pi}(j,s_1) = V^{\pi}(j,s_2)$.

In the Taxi problem, the source and destination of the passenger are
irrelevant to the ${\sf Navigate}(t)$ subtask---only the target $t$
and the current taxi position are relevant.

\condition{Condition 2: Leaf Irrelevance.}   A set of state variables
$Y$ is {\em irrelevant for a primitive action} $a$ if for any pair
of states $s_1$ and $s_2$ that differ only in their values for the
variables in $Y$, \[\sum_{s_1'} P(s_1'|s_1,a) R(s_1'|s_1,a) =
\sum_{s_2'} P(s_2'|s_2,a) R(s_2'|s_2,a).\]

This condition is satisfied by the primitive actions {\sf North}, {\sf
South}, {\sf East}, and {\sf West} in the taxi task, where {\it all}
state variables are irrelevant because $R$ is constant. 

The next two conditions involve ``funnel'' actions---macro actions
that move the environment from some large number of possible states to
a small number of resulting states.  The completion function of such
subtasks can be represented using a number of values proportional to
the number of resulting states.  

\condition{Condition 3: Result Distribution Irrelevance (Undiscounted
case.)}  
A set of state variables $Y_j$ is {\em irrelevant for the result
distribution of action} $j$ if, for all abstract policies $\pi$
executed by $M_j$ and its descendants in the MAXQ hierarchy, the
following holds: for all pairs of states $s_1$ and $s_2$ that differ
only in their values for the state variables in $Y_j$,
\[\lforall{s'}\; P^{\pi}(s'|s_1,j) = P^{\pi}(s'|s_2,j).\]

Consider, for example, the {\sf Get} subroutine under an optimal
policy for the taxi task.  Regardless of the taxi's position in state
$s$, the taxi will be at the passenger's starting location when {\sf
Get} finishes executing (i.e., because the taxi will have just
completed picking up the passenger).  Hence, the taxi's initial
position is irrelevant to its resulting position.  (Note that this is
only true in the undiscounted setting---with discounting, the result
distributions are not the same because the number of steps $N$
required for {\sf Get} to finish depends very much on the starting
location of the taxi.  Hence this form of state abstraction is rarely
useful for cumulative discounted reward.)

\condition{Condition 4: Termination.}  Let $M_j$ be a child task of $M_i$
with the property that whenever $M_j$ terminates, it causes $M_i$ to
terminate too.  Then the completion cost $C(i,s,j) = 0$ and does not
need to be represented.  This is a particular kind of funnel
action---it funnels all states into terminal states for $M_i$. 

For example, in the Taxi task, in all states where the taxi is holding
the passenger, the {\sf Put} subroutine will succeed and result in a
terminal state for {\sf Root}.  This is because the termination
predicate for {\sf Put} (i.e., that the passenger is at his or her
destination location) implies the termination condition for {\sf Root}
(which is the same).  This means that $C({\sf Root}, s, {\sf Put})$ is
uniformly zero, for all states $s$ where {\sf Put} is not terminated.

\condition{Condition 5: Shielding.}  Consider subtask $M_i$ and let
$s$ be a state such that for all paths from the root of the DAG down
to $M_i$, there exists a subtask that is terminated.  Then no $C$
values need to be represented for subtask $M_i$ in state $s$, because
it can never be executed in $s$.

In the Taxi task, a simple example of this arises in the {\sf Put}
task, which is terminated in all states where the passenger is not in
the taxi.  This means that we do not need to represent $C({\sf Root},
s, {\sf Put})$ in these states.  The result is that, when combined
with the Termination condition above, we do not need to explicitly
represent the completion function for {\sf Put} at all!

By applying these abstraction conditions to the Taxi task, the value
function can be represented using 632 values, which is much less than
the 3,000 values required by flat Q learning.  Without state
abstractions, MAXQ requires 14,000 values!

\begin{theorem}  {\bf (Convergence with State Abstraction)}
Let $H$ be a MAXQ task graph that incorporates the five kinds of state
abstractions defined above.  Let $\pi_x$ be an ordered GLIE
exploration policy that is abstract.  Then under the same
conditions as Theorem \ref{theorem-converge}, MAXQ-Q converges with
probability 1 to the unique recursively optimal policy $\pi^*_r$
defined by $\pi_x$ and $H$. 
\end{theorem}

\condition{Proof: (sketch)} Consider a subtask $M_i$ with relevant
variables $X$ and two arbitrary states $(x,y_1)$ and $(x,y_2)$.  We
first show that under the five abstraction conditions, the value
function of $\pi^*_r$ can be represented using $C(i,x,j)$ (i.e.,
ignoring the $y$ values).  To learn the values of $C(i,x,j) =
\sum_{x',N} P(x',N|x,j) V(i,x')$, a Q-learning algorithm needs samples
of $x'$ and $N$ drawn according to $P(x',N|x,j)$.  The second part of
the proof involves showing that regardless of whether we execute $j$
in state $(x,y_1)$ or in $(x,y_2)$, the resulting $x'$ and $N$ will
have the same distribution, and hence, give the correct expectations.
Analogous arguments apply for leaf irrelevance and $V(a,x)$.  The
termination and shielding cases are easy.

\section{Experimental Results}

We implemented MAXQ-Q for a noisy version of the Taxi domain and for
Kaelbling's HDG navigation task \cite{k-hrl:pr-93} using Boltzmann
exploration.  Figure~\ref{fig-compare} shows the performance of flat Q
and MAXQ-Q with and without state abstractions on these tasks.
Learning rates and Boltzmann cooling rates were separately tuned to
optimize the performance of each method.  The results show that
without state abstractions, MAXQ-Q learning is slower to converge than
flat Q learning, but that with state abstraction, it is much faster.

{\footnotesize
\def\normalsize{\footnotesize}
\begin{figure}
\begin{minipage}{2.45in}
{\epsfxsize=2.45in
\centerps{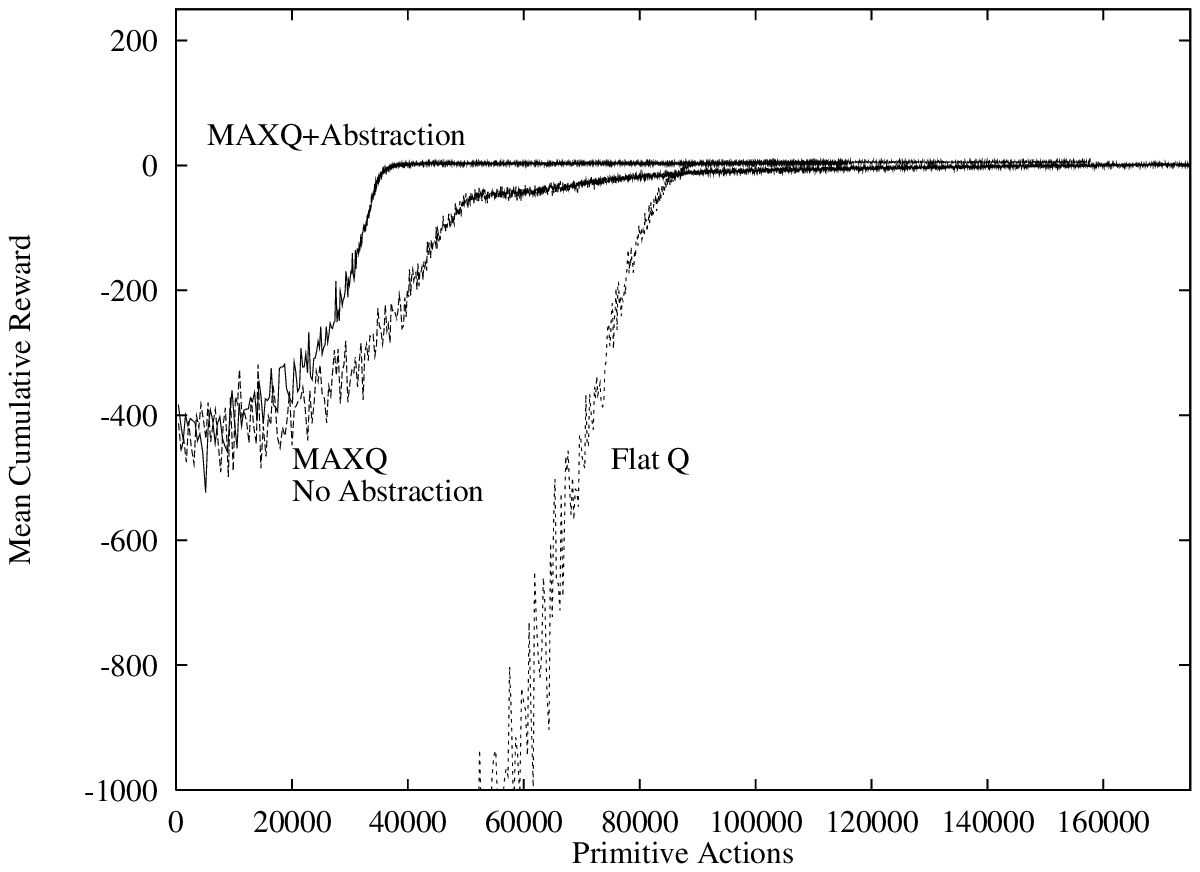}
}
\end{minipage}
\hfil
\begin{minipage}{2.45in}
{\epsfxsize=2.45in
\centerps{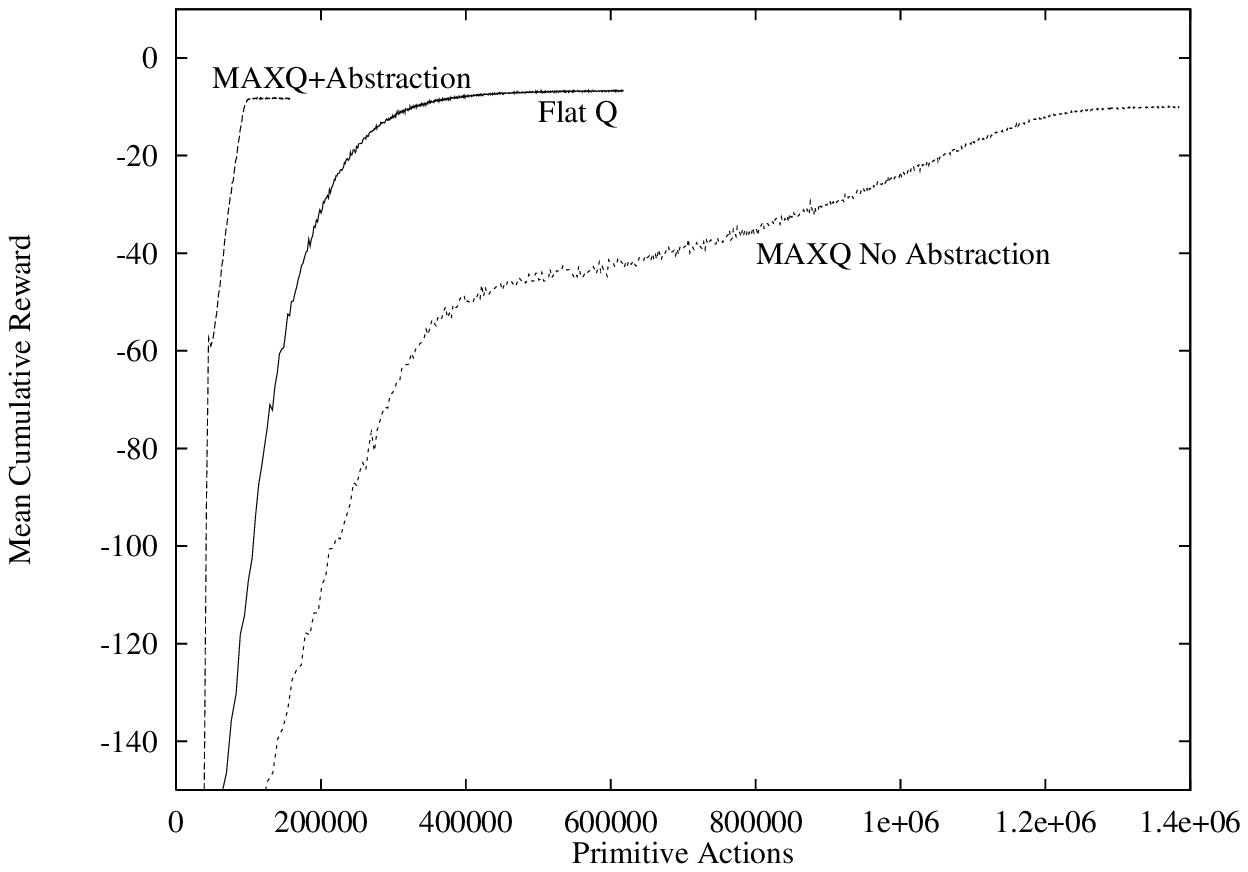}
}
\end{minipage}
\caption{Comparison of MAXQ-Q with and without state abstraction to
flat Q learning on a noisy taxi domain (left) and Kaelbling's HDG task
(right).  The horizontal axis gives the number of primitive actions
executed by each method.  The vertical axis plots the average of 100
separate runs.}
\label{fig-compare}
\end{figure}
}

\section{Conclusion}

This paper has shown that by understanding the reasons that state
variables are irrelevant, we can obtain a simple proof of the
convergence of MAXQ-Q learning under state abstraction.  This is much
more fruitful than previous efforts based only on weak notions of
state aggregation \cite{bt-ndp-96}, and it suggests that future
research should focus on identifying other conditions that permit safe
state abstraction.

\small
\bibliography{c:/tex/bib,c:/tex/colt,c:/tex/nn2,c:/tex/neural-computation,c:/tex/ml,c:/tex/aij,c:/tex/nc,c:/tex/nips-10}
\bibliographystyle{ieeetr}
\end{document}